\documentclass[lettersize,journal]{IEEEtran}
\usepackage{amsmath,amsfonts,amssymb,relsize,bm,breqn,euscript,amsbsy}
\usepackage{algorithm,algorithmic}
\usepackage{array}
\usepackage{textcomp}
\usepackage{stfloats}
\usepackage{url}
\usepackage{verbatim}
\usepackage{graphicx}
\usepackage{balance}

\usepackage{lineno,hyperref}
\usepackage{enumerate}
\usepackage{adjustbox}
\usepackage{multirow}
\usepackage{latexsym}
\usepackage{flushend}
\usepackage{multicol}

\usepackage{booktabs}

\newlength{\bibitemsep}
\newlength{\bibparskip}
\let\oldthebibliography\thebibliography
\renewcommand\thebibliography[1]{%
  \oldthebibliography{#1}%
  \setlength{\parskip}{\bibitemsep}%
  \setlength{\itemsep}{\bibparskip}%
}

\def\+#1{\boldsymbol{#1}}

\def\*#1{\mathbf{#1}}

\newcommand\norm[1]{\left\lVert#1\right\rVert}

\def\BibTeX{{\rm B\kern-.05em{\sc i\kern-.025em b}\kern-.08em
    T\kern-.1667em\lower.7ex\hbox{E}\kern-.125emX}}

\begin{document}
\title{Enhancing Blind Source Separation with Dissociative Principal Component Analysis}
\author{Muhammad Usman Khalid
\thanks{M. U. Khalid is with the College of Computer and Information Sciences, Imam Mohammad Ibn Saud Islamic University, Riyadh 11564, Saudi Arabia. E-mail: mukhalid@imamu.edu.sa}}


\maketitle

\begin{abstract}
Principal component analysis (PCA) and its sparse variants (sPCA) are widely used as a precursor to independent component analysis (ICA) for blind source separation (BSS). However, sPCA typically relies on a deflation strategy that extracts components sequentially and imposes orthogonality between them. When the underlying sources overlap, this discards the cross component structure that ICA depends on, degrading separation. This paper proposes dissociative PCA (DPCA), which estimates components jointly rather than by deflation. DPCA introduces left and right dissociation matrices into the SVD based decomposition to explicitly model the interdependencies among principal components (PCs) and loading vectors (LVs), while sparsity constraints maintain interpretability. We develop three algorithms called DPCA1a, DPCA1b, and DPCA2, using adaptive soft thresholding with gradient and coordinate descent, together with a secondary firm thresholding step that preserves sparsity and suppresses background noise in the recovered loading vectors. The method is evaluated on four settings, namely simulated fMRI source retrieval, foreground and background separation, image reconstruction, and image inpainting, where it recovers source structure more reliably than classical sPCA based pipelines, with the largest gains under significant spatial overlap. DPCA reduces to ordinary PCA when the sparsity parameter is zero. A MATLAB implementation of the proposed algorithms is publicly available at \url{https://github.com/usmankhalid06/DPCA}.
\end{abstract}


\section*{Introduction}
Principal component analysis first introduced in the early twentieth century \cite{pearsonLIII1901}, has been widely employed for data processing, feature extraction, and dimension reduction across various fields, including signal and image processing, machine learning, and exploratory data analysis \cite{kochAnalysis2013}. It projects high-dimensional, correlated data into a lower-dimensional subspace, which is defined by uncorrelated PCs and spanned by orthogonal LVs \cite{hotellingAnalysis1933, stewartEarly1993}. These PCs preserve the majority of the statistical information in the data \cite{jolliffePrincipal2016}. This makes PCA a powerful multivariate statistical technique that can be utilized to reveal true dimensionality of the data \cite{seghouaneSparse2019} and usually deployed as a preprocessing step for BSS algorithms \cite{frimanExploratory2002}.

Principal components, however, are difficult to interpret \cite{jeffersTwo1967} because each of them is a linear combination of all variables from original feature space. To address this shortcoming, several sparse variants of PCA have been proposed \cite{zouSelective2018} that seek to balance interpretability and statistical fidelity by constructing PCs from a minimal number of variables, while retaining as much of the statistical information in the data as possible. On the other hand, sPCA is computationally intractable because it imposes a cardinality constraint on the LVs \cite{tillmannComputational2014a}. To tackle this issue, numerous approximate algorithms have been developed to achieve a solution that is nearly optimal, by striking a balance between computational efficiency and sparsity \cite{wangStatistical2016}.

Sparse PCA is an optimization problem whose objectives are to maximize the variance explained by the PCs while simultaneously achieving sparsity for the LVs \cite{guerra-urzolaGuide2021}. For instance, a first effort at sparse PCA resulted in simplified component technique with LASSO (SCoTLASS) \cite{jolliffeModified2003}, which aimed at maximizing the Rayleigh quotient under $l_1$ norm penalty, a regression-based optimization problem with LASSO penalty led to sparse PCA (SPCA) \cite{zouSparse2006}, a convex relaxation of the sparse PCA problem using semidefinite programming (SDP) produced DSPCA method \cite{daspremontDirect2007}, a framework that considered the covariance matrix for SVD with $l_1$ penalties on left and right singular vectors culminated in penalized matrix decomposition (PMD) \cite{wittenpenalized2009}, lastly, a novel iterative thresholding approach was introduced, effectively estimating principal subspaces and leading eigenvectors, marking further progress in sparse PCA methodologies \cite{maSparse2013}.

Computationally effective techniques for sparse PCA were subsequently developed. Notably, the generalized power method (GPower), which involved maximizing a convex function on a compact set \cite{journeeGeneralized2010}. Later, an effective solution to non-convex sparse eigenvalue problems known as the truncated power method (TPower) was introduced \cite{yuanTruncated2013}. Innovative methods for sparse PCA have emerged in recent years. For instance, a Bayesian approach that generated components with common sparsity patterns \cite{matteiGlobally2016}, an adaptive block sparse PCA method called preserved PCA (PPCA) maintained sparsity across all LVs through group LASSO with adaptive penalization \cite{seghouaneSparse2019}, an efficient online sparse PCA algorithm that combined gradient-based optimization techniques with stochastic gradient descent \cite{qiuGradientbased2023}, and a recent method that improved sparsity by applying orthogonal rotation to PCs before thresholding \cite{chenNew2024a}.

While sPCA improves the interpretability of PCs by enforcing sparsity, it does not necessarily improve source separation. Traditional sPCA relies on a deflation strategy that extracts components one at a time and imposes orthogonality between them. This sequential orthogonalization discards the cross component structure present in the data. When the underlying sources overlap, that discarded structure is precisely what ICA needs to separate them, so deflation based sPCA followed by ICA tends to separate sources poorly \cite{seghouaneSparse2019}. Ordinary PCA, in contrast, lacks interpretability and, when followed by ICA, also struggles to separate sources under significant spatial overlap \cite{khalidUnsupervised2015, seghouaneLearning2016, zhangExperimental2019b}. Both sPCA and ICA and PCA and ICA pipelines are therefore limited for reliable source separation in this regime.

Comparatively little work has addressed the separation of non orthogonal independent components under moderate to significant spatial overlap within the sPCA and ICA setting \cite{zhangExperimental2019b}. This study addresses that gap. Rather than extracting components by deflation, DPCA estimates them jointly and models the interdependencies among PCs and LVs explicitly, aiming to preserve the structure that supports source separation while retaining interpretability through sparsity, particularly under substantial spatial overlap.

The contributions of this work are as follows. First, DPCA models and disentangles the interdependencies within both LVs and PCs, rather than removing them by deflation. Second, three algorithms, DPCA1a, DPCA1b, and DPCA2, are developed, combining adaptive soft thresholding with gradient and coordinate descent to solve the proposed model. Third, a secondary firm thresholding step preserves sparsity and suppresses background noise in the recovered LVs, supporting interpretability and separation together. Finally, the model reduces to ordinary PCA when the sparsity parameter is set to zero, as shown in Proposition 2.

To ensure wide reach and timely impact of this paper, its preprint version is available on arXiv \cite{khalidEnhancing2024}.

\section{Methods and Materials}
\subsection{Background}
Principal component analysis is most effective when a reduced number of components, $K<p$, is sufficient to capture the majority of the variance, thereby providing a simpler, yet insightful, representation of the underlying structure of the data. For this purpose, consider the data matrix $\text{X} \in \mathbb{R}^{n\times p}$ whose each column is mean-centered and with rank $K\le\min(n,p)$, where $\text{X}$ consists of $n$ observations on $p$ variables. In order to find a linear projection of the data that maximizes the variance retained in the lower-dimensional subspace, PCA seeks to solve the following constrained optimization problem that reduces the dimensionality of the data from $p$ to $K$ \cite{jolliffePrincipal2002}
\begin{flalign}
&\arg\max_{\text{w}_i} \ \ \mathrm{var}(\text{X}\text{w}_i) \quad \textrm{sub.to.} \ \ \text{w}_i^\top\text{w}_i =1, \nonumber\\
&\hspace{4.5cm} \text{w}_i^\top\text{w}_j =0 \ \ \mathrm{for}  \ \ i\ne j\nonumber
\end{flalign}
where, $\text{w}_i$ represents the $i$-th principal LV, projection of the data $\text{X}\text{w}_i$ defines the $i$-th PC, and the operator $\mathrm{var}(.)$ refers to the variance of a given variable. The contribution of $\text{X}\text{w}_i$ explains $100\times\lambda_i/\sum_{j=1}^{p} \lambda_j$ percent of the total variance in the data, where $\lambda_i$ is the $i$-th eigenvalue. The solution of the above optimization problem is found in the following decomposition
\begin{flalign}\label{eq1}
&\mathcal{C} = \sum_{i=1}^k\lambda_i\text{w}_i\text{w}_i^\top
\end{flalign}
where $\mathcal{C}= 1/n(\text{X}^\top\text{X})$ represents covariance matrix, and $\lambda_i$ is the $i$-th largest eigenvalue, with $\text{w}_i$ as the corresponding eigenvector. Based on a linear regression approach, an alternative formulation of PCA can be obtained by solving the following least squares problem to estimate the projection matrix $\text{W}$ that contains the $K$ LVs as \cite{zouSparse2006}
\begin{flalign}\label{eq2}
&\text{W} = \arg\min_{\text{W}} \ \ \sum_{j=1}^n \norm{{\text{x}^j}^\top-\text{W}\text{W}^\top{\text{x}^j}^\top}_2^2 \nonumber\\
&\hspace{0.8cm} \textrm{sub.to.} \ \ \text{W}^\top\text{W} =\text{I}_K
\end{flalign}
Usually, PCA is performed through the SVD of the data matrix $\text{X}$, which is expressed as
\begin{flalign}\label{eq3}
\text{X}= \text{UD}\text{W}^\top
\end{flalign}
In this decomposition, the columns in matrix $\text{UD}$ contains the PCs, the matrix $\text{W}$ holds the LVs as its columns, and the diagonal matrix $\text{D} \in \mathbb{R}^{K\times K}$ contains the singular values along the diagonal that is $\text{d}_1 \geq \text{d}_2 \geq \hdots \geq \text{d}_K > 0$, ordered from largest to smallest. Both $\text{U} \in \mathbb{R}^{n\times K}$, which contains the left singular vectors and $\text{W} \in \mathbb{R}^{p\times K}$, which contains the right singular vectors are orthonormal matrices, satisfying $\text{U}^\top\text{U} = \text{W}^\top\text{W} = \text{I}_K$, where $\text{I}_K$ is the identity matrix of size $K\times K$. If equation (\ref{eq1}) can be reformulated as $\mathcal{C} =\text{W}\Lambda\text{W}^\top$ then it is related to singular values in equation (\ref{eq3}) as $\text{D}=|\Lambda|$. Because non-negative singular values in SVD provide a stable and interpretable transformation, in the subsequent section, equation (\ref{eq3}) will be utilized as the basis for introducing the proposed model.

The proposed method distinctly differs from sparse dictionary learning approaches, such as those described in \cite{aharonKSVD2006}, which typically involve learning an overcomplete dictionary where $K >> n$ and a sparse code directly from the data. In contrast, the proposed approach limits the number of components to $K \leq n$. Rather than learning a dictionary and sparse codes anew, it utilizes predefined PCs and LVs to model the data, adapting existing structures to enhance computational efficiency and interpretability.

\subsection{Proposed Model}
This study proceeds under the assumption that each column of the observed data matrix $\text{X}$ is mean-centered, unless stated otherwise. Additionally, the number of PCs, $K$, is predetermined. For methodologies on determining $K$ from the data, reader is referred to this article \cite{chenEstimating2021}, which discusses the use of cross-validated eigenvalues among other techniques. Particulary, for this paper, $K$ is selected based on performance metrics such as peak signal-to-noise ratio (PSNR) and the correlation strength between the ground truth and the reconstructed components.

This section presents the dissociative PCA method designed to effectively disentangle interdependencies within PCs and LVs. The proposed method uses a sequential estimating procedure for both PCs and LVs, in contrast to conventional sparse PCA methods that often use a deflation strategy that may overlook important interdependencies and compromise source separation effectiveness. By employing both coordinate descent and gradient descent techniques, the proposed approach not only enhances the accuracy of source separation but also improves the interpretability of the results.

Consider the matrix $\text{X} \in \mathbb{R}^{n \times p}$, represented in low-rank form as $\text{U}_q \text{D}_q \text{W}_q^\top$, where $K \leq \min(n, p)$. In this representation, $\text{U}_q \in \mathbb{R}^{n \times K}$ and $\text{W}_q \in \mathbb{R}^{p \times K}$ are orthogonal matrices, while $\text{D}_q \in \mathbb{R}^{K \times K}$ is a diagonal matrix of singular values. Left and right dissociative matrices $\Psi \in \mathbb{R}^{K \times K}$ and $\Phi \in \mathbb{R}^{K \times K}$ are introduced here to encapsulate interdependencies within the PCs and LVs, thereby modifying the traditional SVD framework, and its reformulation is expressed as
\begin{flalign}\label{eq4}
\text{X} = \text{U}_q \text{D}_q \text{W}_q^\top = \text{U}_q \Psi \Psi^{-1} \text{D}_q \Phi^{-1} \Phi \text{W}_q^\top = \text{U} \text{V} \text{W}^\top
\end{flalign}
This approach enhances the effectiveness of the model by applying sparsity constraints to $\text{W}$, ensuring that the decomposition $\text{U}\text{V}\text{W}^\top$ accurately represents the underlying data structure. The model leverages the $l_1$ norm to induce sparsity, effectively managing column sparsity. Further details on this implementation are provided in subsequent subsections.

After the estimation of $\Psi$ and $\Phi$, the non-diagonal middle matrix $\text{V}=\Psi^{-1}\text{D}_q\Phi^{-1}$ facilitates understanding of the interactions between PCs and LVs. It is a diagonal matrix for a standard PCA, however, when sparsity is enforced on $\text{W}$, the matrix $\text{V}$ may exhibit small off-diagonal values, indicating interdependencies among the LVs. While DMs allow to break the interdependencies within PCs and within LVs, the weighing matrix $\text{V}$ captures the extent of these interdependencies.

Because the estimation of $\text{U}$ and $\text{W}$ is not contingent upon $\text{V}$, it can be adjusted or even excluded without altering the structural integrity of $\text{U}$ and $\text{W}$. This flexibility allows for an alternative decomposition where $\text{V}$ is omitted. Considering that both $\text{U}\text{W}^\top$ and $\text{U}\text{V}\text{W}^\top$ can serve as approximate estimates of both $\text{X}$ and the traditional $\text{U}_q\text{D}_q\text{W}_q^\top$, hence the following simplified decomposition is proposed
\begin{flalign}\label{eq5}
\text{X}= \text{U}\text{W}^\top
\end{flalign}
This model retains the essential properties of the original decomposition while incorporating dissociation. As discussed in Remark 1, this redistributes the variance structure while a consistent measure of explained variance is retained across all compared methods. \\ \\

\textbf{Remark 1:} The dissociation matrices $\Psi$ and $\Phi$ redistribute the variance across the components relative to the original SVD. In the modified decomposition $\text{U}_q\Psi\Phi\text{W}_q^\top$, the effective middle factor is $\hat{\text{D}}_q=\Psi\Phi$, which in general is not diagonal. Unlike the original diagonal matrix $\text{D}_q$, whose entries $\text{d}_{q,1}\geq\text{d}_{q,2}\geq\hdots\geq\text{d}_{q,K}>0$ are ordered singular values, $\hat{\text{D}}_q$ may contain off-diagonal entries that encode the modeled interdependencies among the PCs and LVs. Because $\Psi^\top\Psi$ and $\Phi^\top\Phi$ need not be identity matrices, the total variance captured by the dissociated decomposition,
\begin{flalign}
\mathrm{Tr}(\hat{\text{D}}_q^\top\hat{\text{D}}_q)= \mathrm{Tr}\big(\Phi^\top\Psi^\top\Psi\Phi\big), \nonumber
\end{flalign}
generally differs from the original $\mathrm{Tr}(\text{D}_q^\top\text{D}_q)$, and typically
\begin{flalign}
\mathrm{Tr}(\hat{\text{D}}_q^\top\hat{\text{D}}_q)\leq \mathrm{Tr}(\text{D}_q^\top\text{D}_q). \nonumber
\end{flalign}
For this reason, and to ensure a consistent variance measure across all algorithms compared in this paper, the percentage of variance explained is computed directly from the reconstruction as $100\times \big[1-{\norm{\text{X}-\text{X}{\text{W}^\top}^{-1}\text{W}^\top}_F^2}/{\norm{\text{X}}_F^2}\big]$ \cite{shenSparse2008, parkCritical2024}, where the inverse of $\text{W}^\top$ is computed using the Moore-Penrose generalized inverse.\\ \\
Utilizing equation (\ref{eq5}) and noting that $\text{W}_q^\top = \text{Z}_q$, an optimization model is formulated that imposes sparsity constraints on the modified loading matrix $\text{Z}$. This model incorporates sparsity on two distinct levels: firstly, on the non-penalized LVs $\text{z}^k = \text{u}_k^\top\text{X}$, and secondly, on the reconstructed LVs $\text{z}^k = \phi^k \text{Z}_q$. The proposed formulation is given by
\begin{flalign}\label{eq6}
&\{\Psi, \Phi, \text{Z}\} = \arg\min_{\Psi, \Phi, \text{Z}}  \|\text{X} - \text{U}_q \Psi \Phi \text{Z}_q\|_F^2 + \sum_{k=1}^{K} \sum_{j=1}^{V} \lambda^k_j |\phi^k \text{z}_{q,j}| \nonumber\\
&\hspace{2cm}\text{sub.to.} \quad \|\text{Z}\|_1 \leq \rho, \quad \|\text{U}_q \psi_k\|_2 = 1
\end{flalign}

In this model, $\lambda_j^k$, derived from the global regularization parameter $\lambda$, is allocated to each entry of $\text{Z}$ as a data-driven parameter. Both $  \lambda$ and $\rho$ serve as regularization parameters that control the trade-off between fitting the model to the data and promoting sparsity. Specifically, $\lambda$ acts as a penalization factor to encourage sparsity, while $\rho$ sets a hard constraint on the $l_1$ norm of components of $\text{Z}$, enforcing a limit on sparsity.

The parameter $\lambda_j^k$, as specified earlier, serves as an adaptive matrix variant of the LASSO penalty \cite{zouAdaptive2006}. This data-driven regularization parameter dynamically adjusts the shrinkage applied to each element of $z_j^k$, allows for applying more substantial shrinkage to entries that are near zero and less to those with significant values, thereby improving threshold outcomes. Conversely, the sparsity constraint $\rho$ employs a firm thresholding technique, which is preferred over hard thresholding due to its non-smooth nature, and over soft thresholding, which may alter significant values. Firm thresholding adeptly maintains the sparsity pattern \cite{voroninNew2013}, ensuring that key features in the reconstructed matrix $\text{Z} = \Phi \text{Z}_q$ are preserved by leaving large coefficients unchanged and removing background noise. Notably, coefficients that fall between two predefined thresholds, $\rho_1$ and $\rho_2$, undergo a controlled shrinkage, which is less severe than in soft thresholding, thus striking an effective balance between maintaining data fidelity and promoting sparsity.

\subsection{Proposed Solution}
A sequential technique is considered to enhance the precision of updating the unknown variables. Initially the error matrix of all signals $\text{E}_{k}$ is considered and given as
\begin{flalign}\label{eq7}
\text{E}_{k} = \text{X}-\sum_{i=1,i\ne k}^{K}\text{u}_i\text{z}^i
\end{flalign}
Subsequently, equation (\ref{eq6}) is reformulated to address a rank-1 minimization problem aimed at updating the dissociation matrices $\Psi/\Phi$, and the reconstructed source matrix $\text{Z}$, element-wise. The updated formulation is as follows
\begin{flalign}\label{eq8}
&\{\psi_k, \phi^k, \text{z}^k\} = \arg\min_{\psi_k, \phi^k, \text{z}^k} \|\text{E}_k - \text{U}_q \psi_k \phi^k \text{Z}_q\|_F^2 + \nonumber\\
&\sum_{j=1}^{V} \lambda^k_{j} |\phi^k \text{z}_{q,j}|, \quad \text{sub.to.} \quad \|\text{z}^k\|_1 \leq \rho, \quad \|\text{U}_q \psi_k\|_2 = 1
\end{flalign}
While considering solving the constraints on $\text{Z}$ as a separate optimization problem, updates for $\psi_k$ and $\phi^k$ are obtained by employing the Lagrange multiplier method. The associated Lagrangian for the optimization problem, defined in equation $(\ref{eq8})$, incorporates the necessary constraints and is expressed as
\begin{flalign}\label{eq9}
\mathcal{L}(\psi_k, \phi^k) = & \|\text{E}_k - \text{U}_q \psi_k \phi^k \text{Z}_q\|_F^2 + \sum_{j=1}^{V} \lambda^k_j |\phi^k \text{z}_{q,j}| +\nonumber\\
&\hspace{3.5cm} \mu \left(\|\text{U}_q \psi_k\|_2^2 - 1\right)
\end{flalign}
where $\mu$ is the Lagrange multiplier. This equation can also be expressed as
\begin{flalign}\label{eq10}
&\mathcal{L}(\psi_k,\phi^k)=\mathrm{tr}(\text{E}_k^\top\text{E}_k)-2\mathrm{tr}(\text{E}_k^\top\text{U}_q\psi_{k}\phi^k\text{Z}_q)+\mathrm{tr}(\text{Z}_q^\top{\phi^k}^\top\psi_k^\top\nonumber\\
&\hspace{1.8cm}\text{U}_q^\top\text{U}_q\psi_{k}\phi^k\text{Z}_q)+\sum_{j=1}^{V}\lambda^k_{j}|\phi^k\text{z}_{q,j}|+\mu (\|\text{U}_q \psi_k\|_2^2 - 1)
\end{flalign}
According to Karush-Kuhn-Tucker (KKT) condition for the optimal solution, $\frac{\partial \mathcal{L}}{\partial \phi^k}=0$ is considered to obtain
\begin{flalign}\label{eq11}
&-2\psi_{k}^\top\text{U}_q^\top\text{E}_{k}\text{Z}_q^\top+2\psi_{k}^\top\text{U}_q^\top\text{U}_q\psi_{k}\phi^k\text{Z}_q\text{Z}_q^\top+\nonumber\\
&\hspace{3.8cm}\sum_{j=1}^{V}\lambda^k_{j}\frac{\partial|\phi^k\text{z}_{q,j}|}{\partial\phi^k}=0
\end{flalign}
After some manipulations, the above equation can be addressed using an adaptive LASSO method, where the regularization parameter \(\lambda^k_j\) is defined as \(\lambda/|\text{z}^k_j|\) to apply more substantial shrinkage to smaller coefficients \cite{seghouaneConsistent2018}. The resulting soft thresholding based closed form solution to the above equation is obtained as
\begin{flalign}\label{eq12}
&\phi^k=\mathrm{sgn}\Big(\text{u}_{k}^\top \text{E}_{k}\Big)\circ\Big(|\text{u}_{k}^\top \text{E}_{k}|-\frac{\lambda1}{2|\text{u}_{k}^\top \text{E}_{k}|}\Big)_+\text{Z}_q^\top\Big(\text{Z}_q\text{Z}_q^\top\Big)^{-1}
\end{flalign}
where $\mathcal{T}_\upsilon(\text{y})=\mathrm{sgn}(\text{y})\circ(|\text{y}|-\upsilon1/|\text{y}|)_+$, $(\mathrm{y})_+$, $\mathrm{sgn}(.)$, and $\circ$ deﬁne the component-wise max between $(0,\mathrm{y})$, the component-wise sign, and the Hadamard product, respectively \cite{fuPenalized1998}, and $1$ is a vector of ones. Subsequently, to enforce the $l_1$ constraints on $\text{Z}$, the following optimization problem is considered
\begin{flalign}\label{eq13}
&{\text{z}^k}=\arg \min_{\text{z}^k}\norm{\text{z}^k-\phi^k\text{Z}_q} _2^2 \quad \textrm{sub.to.} \  \|\text{z}^k\|_1 \leq \rho
\end{flalign}
This problem was solved using firm thresholding as
\begin{flalign}\label{eq14}
&\text{z}^k = \mathcal{T}_{\rho_1, \rho_2}(\phi^k\text{Z}_q),
\end{flalign}
where
\[
\mathcal{T}_{\rho_1, \rho_2}(y) =
\begin{cases}
0 & \text{if } |y| \leq \rho_1\\
\frac{\rho_2 (|y| - \rho_1)}{\rho_2 - \rho_1} \mathrm{sgn}(y) & \text{if } \rho_1 < |y| < \rho_2 \\
y & \text{if } |y| \geq \rho_2
\end{cases}
\]
To obtain the closed form solution for $\psi_k$, one can refer to equation (\ref{eq10}), and apply the KKT conditions by computing the gradient of the Lagrangian $\mathcal{L}(\psi_k,\phi^k)$ with respect to $\psi_k$  and setting it to zero $ \frac{\partial \mathcal{L}}{\partial \psi_k} = 0 $. This condition helps derive the necessary expressions to solve for $\psi_k$ under the constraints specified in the optimization model
\begin{flalign}
&-2\text{U}_q^\top\text{E}_{k}\text{Z}_q^\top\phi^{k^\top}+2\text{U}_q^\top\text{U}_q\psi_{k}\phi^k\text{Z}_q\text{Z}_q^\top\phi^{k^\top}+2\mu\text{U}_q^\top\text{U}_q\psi_{k}=0\nonumber
\end{flalign}
which leads to
\begin{flalign}\label{eq15}
\psi_{k}=(\text{U}_q^\top\text{U}_q)^{-1}\text{U}_q^\top\text{E}_k\text{Z}_{q}^\top{\phi^k}^\top
\end{flalign}
where $\mu+\phi^k\text{Z}_q\text{Z}_q^\top{\phi^k}^\top$ appeared as the normalization factor in the above equation, but as dictionary columns are normalized to one, this factor can be disregarded. The following algorithms are now described to solve the optimization model given in equation (\ref{eq6}) using gradient descent and coordinate descent, respectively.
\subsection{Proposed Algorithms}
\subsubsection{DPCA1}
In this section, two variants of the gradient descent approach are introduced. The first variant, designated as DPCA1a, is computationally efficient but may exhibit convergence instabilities. The second variant, named DPCA1b, while stable, incurs a higher numerical burden. Besides, DPCA1a retains more variance from the observed data by not imposing second-level sparsity constraints on the reconstructed loading vector $\text{Z}$. However, this approach may result in the inclusion of background noise within each loading vector.
\paragraph{DPCA1a:}
For DPCA1a, an alternative optimization model is used, which is given as follows
\begin{flalign}\label{eq16}
&\{\psi_k, \phi^k\} = \arg\min_{\psi_k, \phi^k, \text{z}^k} \|\text{X} - \text{U}_q \psi_k \phi^k \text{Z}_q\|_F^2 + \sum_{j=1}^{V} \lambda^k_{j} |\phi^k \text{z}_{q,j}|, \nonumber\\
&\hspace{1.7cm} \text{sub.to.} \quad \|\text{U}_q \psi_k\|_2 = 1
\end{flalign}
DPCA1a adopts a gradient descent approach, offering a novel method for computing $\psi_k$ and $\phi_k$. Initially, the Lagrangian is formulated for the respective equation and the KKT conditions are applied. This procedure enables to derive the solution for $\phi^k$ using soft thresholding, as demonstrated in equation (\ref{eq12}) and given with respect to equation (\ref{eq16}) as
\begin{table}[H]
\caption{Algorithm for solving the minimization problem (\ref{eq16})}
\begin{tabular}{ l p{8.2cm}}
\toprule
\multicolumn{2}{p{8.5cm}}{\textbf{Given}: $\text{X} \in \mathbb{R}^{n\times p}$, $\text{U}_q \in \mathbb{R}^{n\times K}$ $\text{Z}_q \in \mathbb{R}^{K\times p}$, $\lambda$, $L$, $K$ } \\
\hline
          1. \rule{0pt}{0ex} \textbf{Initialize} \\
          \rule{12pt}{0ex} $\Psi\leftarrow\text{I}_{K\times K}$, $\Phi\leftarrow0$, $\text{U}\leftarrow\text{U}_q\Psi$, $\text{X}\leftarrow\Phi\text{X}_q$, $l\leftarrow0$\\
          2. \rule{0pt}{0ex} \textbf{Repeat while} $\norm{\text{U}-\text{U}_l}_F/\norm{\text{U}_l}_F>0.01$ \\
          \rule{20pt}{0ex} $l\leftarrow l+1$, $\text{U}_l\leftarrow \text{U}$ \\
          3. \rule{11pt}{0ex} \underline{\textbf{Update} $\phi^k$}\\
          \rule{21pt}{0ex} \textbf{for} $k\leftarrow 1:K$ \\
          \rule{32pt}{0ex} $\phi^k\leftarrow\mathrm{sgn}\Big(\text{u}_{k}^\top\text{X}\Big)\circ\Big(|\text{u}_{k}^\top\text{X}|-\frac{\lambda1}{2|\text{u}_{k}^\top\text{X}|}\Big)_+\text{Z}_q^\top
          \Big(\text{Z}_q\text{Z}_q^\top\Big)^{-1}$ \\
          \rule{32pt}{0ex} $\text{z}^k\leftarrow\phi^k\text{Z}_q$\\
          \rule{21pt}{0ex} \textbf{end}\\
          4. \rule{11pt}{0ex} $\text{A} \leftarrow \text{Z}\text{Z}^\top$ \\
          5. \rule{11pt}{0ex} $\text{B} \leftarrow \text{X}\text{Z}^\top$ \\
          6. \rule{11pt}{0ex} \underline{\textbf{Update} $\psi_k$}\\
          \rule{21pt}{0ex} \textbf{Repeat while} $\norm{\text{U}-\text{U}_o}_F>10^{-5}$ \\
          \rule{32pt}{0ex} \textbf{for} $k\leftarrow 1:K$ \\
          \rule{43pt}{0ex} $\psi_{k} \leftarrow (\text{U}_q^\top \text{U}_q)^{-1} \text{U}_q^\top \left(\frac{1}{a_{k}^{k}} (\text{b}_k - \text{U} a_k) + \text{u}_k\right)$ \\
          \rule{43pt}{0ex} $\psi_k\leftarrow\psi_k/\mathrm{max}(\norm{\text{U}_q\psi_k}_2,1)$\\
          \rule{43pt}{0ex} $\text{u}_k\leftarrow\text{U}_q\psi_{k}$\\
          \rule{32pt}{0ex} \textbf{end}\\
          \rule{32pt}{0ex} $\text{U}_0\leftarrow\text{U}$\\
          \rule{22pt}{0ex} \textbf{end} \\
          7. \rule{0pt}{0ex} \textbf{end} \\
\hline
\rule{0pt}{0ex} \textbf{Output}: $\text{U} $ and $\text{Z}$\\
\hline
\end{tabular}
\label{T1}
\end{table}
\begin{flalign}\label{eq17}
&\phi^k=\mathrm{sgn}\Big(\text{u}_{k}^\top \text{X}\Big)\circ\Big(|\text{u}_{k}^\top \text{X}|-\frac{\lambda1}{2|\text{u}_{k}^\top \text{X}|}\Big)_+\text{Z}_q^\top\Big(\text{Z}_q\text{Z}_q^\top\Big)^{-1}
\end{flalign}
This update is systematically applied across all rows of $\phi^k$. Instead of relying on the computationally intensive method of using the error matrix for updating each column of $\psi_k$, an approach inspired by online learning \cite{mairalOnline2009} is adopted. One can construct loading vectors' correlation profiles and temporal profiles, specifically $\text{A} = \text{Z}\text{Z}^\top$ and $\text{B} = \text{X}\text{Z}^\top$, to facilitate the updates. Utilizing these profiles, each column of $\psi_k$ can be updated efficiently as follows
\begin{flalign}\label{eq18}
\psi_{k} = (\text{U}_q^\top \text{U}_q)^{-1} \text{U}_q^\top \left(\frac{1}{a_{k}^{k}} (\text{b}_k - \text{U} a_k) + \text{u}_k\right)
\end{flalign}
The equation (\ref{eq18}) is applied iteratively until convergence is achieved. To enhance computational efficiency, the right dissociation matrix $k$-th row $\phi^{k}$ is updated only once, rather than being repeatedly until convergence. These update routines are presented in Table 1 in the form of an algorithm given the training data $\text{X}$, PCs/LVs $\text{U}_q/\text{Z}_q$, tuning parameter $\lambda$, number of algorithm iterations $L$, and number of components to be recovered $K$.

\paragraph{DPCA1b:}
For DPCA1b, the optimization model described in equation (\ref{eq8}) is utilized but after replacing the error matrix with the data matrix as
\begin{flalign}\label{eq19}
&\{\psi_k, \phi^k, \text{z}^k\} = \arg\min_{\psi_k, \phi^k, \text{z}^k} \|\text{X} - \text{U}_q \psi_k \phi^k \text{Z}_q\|_F^2 + \sum_{j=1}^{V} \lambda^k_{j} |\phi^k \text{z}_{q,j}|, \nonumber \\ &\hspace{2.3cm}\text{sub.to.} \quad \|\text{z}^k\|_1 \leq \rho, \quad \|\text{U}_q \psi_k\|_2 = 1
\end{flalign}
\begin{table}[H]
\caption{Algorithm for solving the minimization problem (\ref{eq19})}
\begin{tabular}{ l p{8.2cm}}
\toprule
\multicolumn{2}{p{8.5cm}}{\textbf{Given}: $\text{X} \in \mathbb{R}^{n\times p}$, $\text{U}_q \in \mathbb{R}^{n\times K}$ $\text{Z}_q \in \mathbb{R}^{K\times p}$, $\lambda, \rho_1, \rho_2$, $L$, $K$} \\
\hline
          1. \rule{0pt}{0ex} \textbf{Initialize} \\
          \rule{12pt}{0ex} $\Psi\leftarrow\text{I}_{K\times K}$, $\Phi\leftarrow0$, $\text{U}\leftarrow\text{U}_q\Psi$, $\text{X}\leftarrow\Phi\text{X}_q$, $l\leftarrow0$\\
          2. \rule{0pt}{0ex} \textbf{Repeat while} $\norm{\text{U}-\text{U}_l}_F/\norm{\text{U}_l}_F>0.01$ \\
          \rule{20pt}{0ex} $l\leftarrow l+1$, $\text{U}_l\leftarrow \text{U}$ \\
          3. \rule{11pt}{0ex} $\text{A}\leftarrow\text{U}^\top\text{U}$ \\
          4. \rule{11pt}{0ex} $\text{B}\leftarrow\text{U}^\top\text{X}$ \\
          5. \rule{11pt}{0ex} \underline{\textbf{Update} $\phi^k$}\\
          \rule{21pt}{0ex} \textbf{Repeat while} $\norm{\text{Z}-\text{Z}_o}_F>10^{-5}$ \\
          \rule{32pt}{0ex} \textbf{for} $k\leftarrow 1:K$ \\
          \rule{43pt}{0ex} $y\leftarrow\frac{1}{a_{k}^{k}}\Big(\text{b}^k - a^k\text{Z}\Big)+ \text{z}^k$ \\
          \rule{43pt}{0ex} $\phi^k\leftarrow\mathrm{sgn}(y)\circ\Big(|y|-\frac{\lambda1}{2|y|}\Big)_+\text{Z}_q^\top\Big(\text{Z}_q\text{Z}_q^\top\Big)^{-1}$ \\
          \rule{43pt}{0ex} $\text{z}^k\leftarrow\phi^k\text{Z}_q$\\
          \rule{32pt}{0ex} \textbf{end}\\
          \rule{32pt}{0ex} $\text{Z}_0\leftarrow\text{Z}$\\
          \rule{22pt}{0ex} \textbf{end} \\
          6. \rule{11pt}{0ex} \underline{\textbf{Update} $z^k$}\\
          \rule{21pt}{0ex}  Use equation \ref{eq13} to update $\text{z}^k$, \quad for \quad $k=\{1,\hdots,K\}$\\
          7. \rule{11pt}{0ex} $\text{A} \leftarrow \text{Z}\text{Z}^\top$ \\
          8. \rule{11pt}{0ex} $\text{B} \leftarrow \text{X}\text{Z}^\top$ \\
          9. \rule{11pt}{0ex} \underline{\textbf{Update} $\psi_k$}\\
          \rule{21pt}{0ex} \textbf{Repeat while} $\norm{\text{U}-\text{U}_o}_F>10^{-5}$ \\
          \rule{32pt}{0ex} \textbf{for} $k\leftarrow 1:K$ \\
          \rule{43pt}{0ex} $\psi_{k} \leftarrow(\text{U}_q^\top \text{U}_q)^{-1} \text{U}_q^\top \left(\frac{1}{a_{k}^{k}} (\text{b}_k - \text{U} a_k) + \text{u}_k\right)$ \\
          \rule{43pt}{0ex} $\psi_k\leftarrow\psi_k/\mathrm{max}(\norm{\text{U}_q\psi_k}_2,1)$\\
          \rule{43pt}{0ex} $\text{u}_k\leftarrow\text{U}_q\psi_{k}$\\
          \rule{32pt}{0ex} \textbf{end}\\
          \rule{32pt}{0ex} $\text{U}_0\leftarrow\text{U}$\\
          \rule{22pt}{0ex} \textbf{end} \\
          10. \rule{0pt}{0ex} \textbf{end} \\
\hline
          \rule{0pt}{0ex} \textbf{Output}: $\text{U} $ and $\text{Z}$\\
\hline
\end{tabular}
\label{T2}
\end{table}

This algorithm not only optimizes $\psi_k$ but also refines the computation of $\phi^{k}$, applying it repeatedly until convergence is achieved. It leverages the principal components' correlation profiles ($\text{A}=\text{U}^\top\text{U}$) and spatial profiles ($\text{B}=\text{U}^\top\text{X}$) for reaching computationally efficient updates. Because the subproblems for $\psi_k$ and $\phi^k$ are each solved in closed form, this approach produces a stable update for both, with the objective in equation (\ref{eq8}) decreasing monotonically across the alternating updates. Utilizing the online approach \cite{mairalOnline2009}, the update for $\phi^k$ can be obtained using the gradient descent approach, soft thresholding, and least squares as

\begin{flalign}\label{eq20}
&\phi^k=\mathrm{sgn}\Big(\frac{1}{a_{k}^{k}} (\text{b}^k - a^k\text{Z}) + \text{z}^k\Big)\circ\Big(|\frac{1}{a_{k}^{k}} (\text{b}^k - a^k\text{Z}) + \text{z}^k|\nonumber\\ &\hspace{1.2cm}-\frac{\lambda1}{2|\frac{1}{a_{k}^{k}} (\text{b}^k - a^k\text{Z}) + \text{z}^k|}\Big)_+\text{Z}_q^\top\Big(\text{Z}_q\text{Z}_q^\top\Big)^{-1}
\end{flalign}
This is followed by the second-level firm thresholding on reconstructed LVs as given by equation (\ref{eq13}) and iterative updates of modified PCs by equation (\ref{eq18}). These update procedures are outlined as an algorithm in Table 2.

\subsubsection{DPCA2}
Coordinate descent's approach of updating LVs and its associated PC elements as pair effectively leverages and preserves the relationship between these elements, often resulting in faster convergence and improved performance. This approach is particularly advantageous in scenarios characterized by high-dimensional data with inherent structures or sparsity.
\begin{table}[H]
\caption{Algorithm for solving the minimization problem (\ref{eq8})}
\begin{tabular}{ l p{8.2cm}}
\toprule
\multicolumn{2}{p{8.5cm}}{\textbf{Given}: $\text{X} \in \mathbb{R}^{n\times p}$, $\text{U}_q \in \mathbb{R}^{n\times K}$ $\text{Z}_q \in \mathbb{R}^{K\times p}$, $\lambda, \rho_1, \rho_2$, $L$, $K$ } \\
\hline
          1. \rule{0pt}{0ex} \textbf{Initialize} \\
          \rule{12pt}{0ex} $\Psi\leftarrow\text{I}_{K\times K}$, $\Phi\leftarrow0$, $\text{U}\leftarrow\text{U}_q\Psi$, $\text{X}\leftarrow\Phi\text{X}_q$, $l\leftarrow0$\\
          2. \rule{0pt}{0ex} \textbf{Repeat while} $\norm{\text{U}-\text{U}_l}_F/\norm{\text{U}_l}_F>0.01$ \\
          \rule{20pt}{0ex} $l\leftarrow l+1$, $\text{U}_l\leftarrow \text{U}$ \\
          3. \rule{12pt}{0ex}  \textbf{for} $k \leftarrow 1 : K$ \\
          4. \rule{21pt}{0ex} \underline{\textbf{Compute}}\\
          \rule{32pt}{0ex} $\psi_k\leftarrow 0$, $\phi^k\leftarrow 0$, $\text{z}^k\leftarrow 0$ \\
          \rule{32pt}{0ex} $\text{E}_k\leftarrow\text{X} - \text{U}\text{Z}$\\
          \rule{32pt}{0ex} $\text{z}^k \leftarrow \text{u}_{k}^\top\text{E}_{k}$\\
          5. \rule{21pt}{0ex} \underline{\textbf{Update} $\phi^k$}\\
          \rule{32pt}{0ex} $\phi^k\leftarrow\mathrm{sgn}\Big(\text{z}^{k}\Big)\circ\Big(|\text{z}^{k}|-\frac{\lambda1}{2|\text{z}^{k}|}\Big)_+\text{Z}_q^\top\Big(\text{Z}_q\text{Z}_q^\top\Big)^{-1}$\\
          \rule{32pt}{0ex} $\text{z}^k\leftarrow\phi^k\text{Z}_q$\\
          6. \rule{21pt}{0ex} \underline{\textbf{Update} $z^k$}\\
          \rule{32pt}{0ex}  Use equation \ref{eq13} to update $\text{z}^k$\\
          7. \rule{21pt}{0ex} \underline{\textbf{Update} $\psi_k$}\\
          \rule{32pt}{0ex} $\psi_{k}\leftarrow(\text{U}_q^\top\text{U}_q)^{-1}\text{U}_q^\top\text{E}_k{\text{z}^k}^\top$ \\
          \rule{32pt}{0ex} $\psi_{k} \leftarrow \psi_{k} /\norm{\text{U}_q\psi_{k}}_2$ \\
          \rule{32pt}{0ex} $\text{u}_k\leftarrow\text{U}_q\psi_{k}$\\
          8. \rule{12pt}{0ex} \textbf{end} \\
          9. \rule{0pt}{0ex} \textbf{end} \\
\hline
          \rule{0pt}{0ex} \textbf{Output}: $\text{U} $ and $\text{Z}$\\
\hline
\end{tabular}
\label{T3}
\end{table}
Therefore, coordinate descent is employed to solve the proposed cost function in equation (\ref{eq8}), where the $k$-th column and row of $\Psi$ and $\Phi$ respectively are updated as a pair. This process is repeated until all $K$ elements have been updated.  The update for $\psi_k$ is obtained using equation (\ref{eq15}) where as the update of $\phi_k$ and $\text{z}^k$ are obtained using equation (\ref{eq12}) and (\ref{eq13}), respectively. The algorithm facilitating these update processes is detailed in Table 3. \\ \\
\textbf{Proposition 2:} The proposed model in equation (\ref{eq16}) adapts seamlessly to function as ordinary PCA when the sparsity parameter is reduced to zero, which leads to $\text{V}_1=\text{U}_q^{-1}\text{X}\text{Z}_q^{-1}$, otherwise $\text{V}_2=\text{V}_1\Phi^{-1}\Psi^{-1}\text{V}_1$.\\
\textbf{Proof:} Assume that for ordinary PCA, data matrix is decomposed as $\text{X}=\text{U}_q\text{V}_1\text{Z}_q$ and for the proposed PCA it is $\text{X}=\text{U}_q\Psi\Phi\text{Z}_q$ then this implies that $\text{V}_1 = \text{V}_2$ where $\text{V}_2=\Psi\Phi$ only when $\lambda =0$, thus the proposed model reduces to ordinary PCA. This is because $\Psi$ reduces to identity matrix and $\Phi$ becomes $\text{V}_1$. In contrast, when $\lambda>0$ then given that $\text{V}_2 = \Psi\Phi $ and $\text{X}=\text{U}_q\Psi\Phi\text{Z}_q$, $\Psi$ and $\Phi$ are obtained as
\begin{flalign}
&\Psi= \text{U}_q^{-1}\text{X}\text{Z}_q^{-1} \Phi^{-1} \nonumber\\
&\Phi= \Psi^{-1}\text{U}_q^{-1}\text{X}\text{Z}_q^{-1}  \nonumber
\end{flalign}
This implies that $\text{V}_2 = \text{U}_q^{-1}\text{X}\text{Z}_q^{-1} \Phi^{-1}\Psi^{-1}\text{U}_q^{-1}\text{X}\text{Z}_q^{-1}  =\text{V}_1\Phi^{-1}\Psi^{-1}\text{V}_1$. This completes the proof.

\section{Experiments}
This section evaluates the performance of the proposed algorithms DPCA1a, DPCA1b, and DPCA2, and compares them against established sparse PCA and ICA algorithms across four distinct image applications. The comparison includes the following algorithms: fastICA\footnote{\url{https://research.ics.aalto.fi/ica/fastica/}}, SPCA\footnote{\url{https://www.jstatsoft.org/article/view/v084i10}}, PMD\footnote{\url{https://github.com/idnavid/sparse_PCA/tree/master/code}}, PPCA\footnote{\url{https://github.com/idnavid/sparse_PCA}}, and GPower\footnote{\url{http://www.montefiore.ulg.ac.be/~journee/GPower.zip}}. MATLAB source code for each algorithm is accessible through the provided links.

The evaluation consists of four cases, which is a medical imaging based simulation study, along with three image-based applications including background subtraction, image reconstruction, and image inpainting. In each case, the performance of all algorithms is compared based on their effectiveness in retrieving the ground truth from the observed data.

\begin{figure*}[t]
\includegraphics[width=14cm]{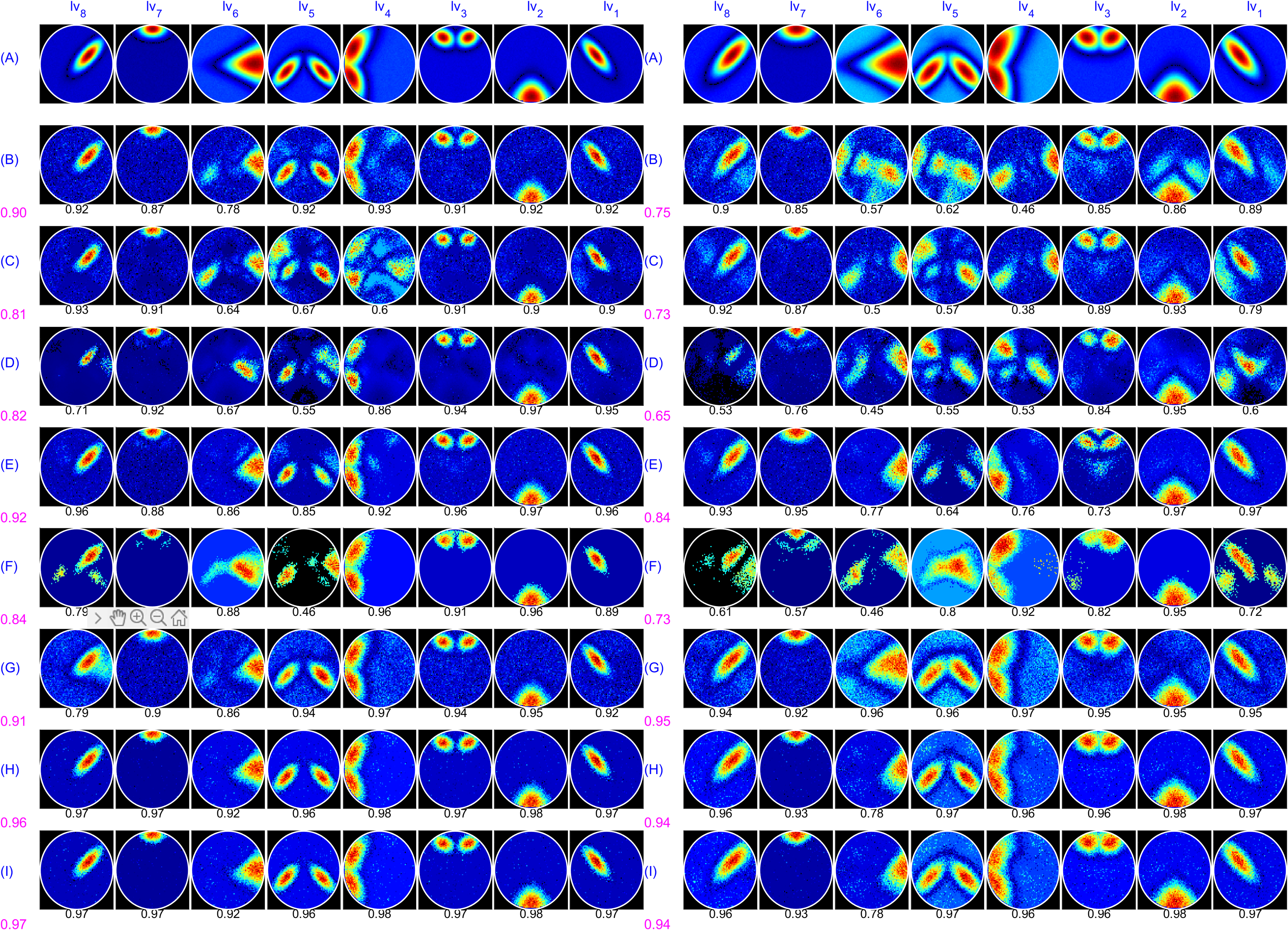}
\centering
\caption{Row (A) displays the ground truth spatial maps with moderate (left, $\rho=6$) and significant (right, $\rho=12$) overlaps. Rows (B) through (I) show the maps recovered as loading vectors/sources by various algorithms including B) PCA+ICA, C) PPCA+ICA, D) SPCA+ICA, E) PMD, F) GPower, G) DPCA1a, H) DPCA1b, and I) DPCA2. Correlation values between the ground truth and the recovered sources are indicated for each map, with the mean correlation displayed on the left.\label{f1}}
\vspace{-3mm}
\end{figure*}

\subsection{Simulation study}
\subsubsection{Synthetic dataset generation}
The Simtb toolbox \cite{erhardtSimTB2011} was utilized to create a synthetic dataset that mimics the characteristics of experimental fMRI signals, which are distinguished by their sparse characteristics \cite{seghouaneHierarchical2012}. Using this toolbox, eight distinct PCs and LVs were created. Each PC consisted of $240$ data points that were taken from discrete cosine transform with bases numbers $\{3, 11, 19, 27, 35, 43, 51, 59\}$, and each loading vector was represented by a $70\times70$ voxel square image. All PCs were normalized to have zero mean and a variance of one. The LVs were created using these source identifiers $\{30, 6, 2, 20, 9, 23, 11, 29\}$. Image slice activations were produced by deploying Gaussian distributions defined by parameters for position, orientation, and spread. The spatial coverage of these activations was regulated through the spread parameter ($\rho$), represented by Gaussian distribution as $\sim \mathcal{N}(\rho,\,0.05)$.

The LVs with two different i) moderate (left), and ii) significant (right) spatial overlaps are shown in Figure $\ref{f1}$A, which are also considered as the ground truth (GT) LVs when retrieving the underlying spatial sources. The GT-PCs and LVs were used to construct a synthetic data matrix through a linear mixture model defined as $\text{X} = \sum_{i=1}^{8}(\text{pc}_i + \omega_i)(\text{lv}^i + \gamma^i)$. In this model, the noise matrices $\Omega \in \mathbb{R}^{240 \times 8}$ and $\Gamma \in \mathbb{R}^{8 \times 4900}$ were generated from Gaussian distributions, specifically $\Omega \sim \mathcal{N}(0,\,\eta_t)$ and $\Gamma \sim \mathcal{N}(0,\,\eta_s)$, where $\eta_t$ and $\eta_s$ are temporal and spatial noise variances, respectively. The matrix $\text{PC} \in \mathbb{R}^{240 \times 8}$ represents the time courses, and $\text{LV} \in \mathbb{R}^{8 \times 4900}$ contains the spatial maps obtained by reshaping the image slices. The synthetic dataset $\text{X}$ was created based on the spread parameter $\rho$, spatial noise $\eta_s$, and temporal noise $\eta_t$, and subsequently used for source retrieval across all participating algorithms.

\subsubsection{Synthetic dataset configuration and parameters}
Various algorithms were assessed to understand their effectiveness in data recovery. For instance, traditional PCA combined with spatial ICA (PCA+ICA) did not enhance interpretability nor did it effectively recover underlying sources, particularly in cases with considerable spatial overlaps. While PPCA+ICA marginally improved interpretability by leveraging sparse components, it was still inadequate for robust source separation. SPCA integrated with ICA (SPCA+ICA) also underperformed, particularly with significant spatial overlaps. These observations underscore the limitations of combining PCA with ICA and highlight the necessity for the proposed algorithms, which aim to simultaneously boost interpretability and ensure effective source separation.

For all algorithms, eight components were extracted. The sparsity parameter for each algorithm was selected based on optimal performance measures, such as the correlation between retrieved components and generated sources. For PPCA, the sparsity parameter was set to $0.005$. The sparsity parameter for SPCA was chosen according to the ground truth, which corresponds to the number of active pixels in the source slices. For PMD, the sparsity parameter was set to $0.43$, and for GPower, it was determined to be $0.3$. For the proposed algorithms, the sparsity parameter was normalized using a factor based on the data matrix dimensions, specifically $\sqrt{n \times p}$. This resulted in a sparsity parameter of $\lambda = 0.67$ for all proposed algorithms, with $\rho_1 = 0.12$ and $\rho_2 = 0.24$ for DPCA1b and DPCA2. The proposed algorithms and PPCA were run for $30$ iterations, while the default iteration settings were used for all other algorithms.

\begin{table*}[t]
\centering
\caption{The average correlation values for the eight loading vectors ($lv_1$–$lv_8$) over 15 trials are presented, along with the mean and standard deviation (Std) for each algorithm. The highest and second-highest values are highlighted in bold.}
\begin{tabular}{l| c c c c c c c c |c| c}\toprule
Algos   & $lv_1$  & $lv_2$  & $lv_3$  & $lv_4$ & $lv_5$  & $lv_6 $ & $lv_7$   & $lv_8$  & Mean & Std   \\\midrule
PCA+ICA  & 0.909 & 0.932 & 0.919 & 0.779 & 0.810 & 0.720 & 0.870 & 0.919 & 0.857 & 0.048\\
PPCA+ICA & 0.899 & 0.914 & 0.922 & 0.608 & 0.696 & 0.649 & 0.906 & 0.930 & 0.816 & 0.028\\
SPCA+ICA & 0.916 & 0.960 & 0.924 & 0.846 & 0.582 & 0.598 & 0.913 & 0.763 & 0.813 & 0.037\\
PMD      & 0.949 & 0.970 & 0.930 & 0.911 & 0.887 & \bf0.850 & 0.914 & 0.947 & 0.920 & 0.026\\
GPower   & 0.884 & 0.957 & 0.898 & 0.951 & 0.601 & \bf0.847 & 0.856 & 0.824 & 0.852 & 0.056\\
DPCA1a   & 0.926 & 0.947 & 0.942 & 0.967 & 0.879 & 0.819 & 0.893 & 0.867 & 0.905 & 0.032\\
DPCA1b   & \bf0.967 & \bf0.978 & \bf0.972 & \bf0.979 & \bf0.911 & 0.799 & \bf0.966 & \bf0.967 & \bf0.942 & \bf0.019\\
DPCA2    & \bf0.967 & \bf0.978 & \bf0.973 & \bf0.979 & \bf0.899 & 0.787 & \bf0.965 & \bf0.968 & \bf0.939 & \bf0.020\\
\bottomrule
\end{tabular}
\label{T4}
\end{table*}

\subsubsection{Synthetic dataset results}
The results in Figure $\ref{f1}$ demonstrate the effectiveness of eight algorithms (B to I) in recovering ground truth spatial maps (LVs) from a generated dataset $\text{X}$. The top row (A) shows the ground truth LVs under two conditions: moderate spatial overlap on the left and significant spatial overlap on the right. This ground truth, along with setting $\eta_t = 0.9$ and $\eta_s = 0.005$, was used to generate the data matrix $\text{X}$ according to the discussion in the previous subsection. Each recovered map has an associated correlation value, reflecting its similarity to the ground truth. Higher correlation values indicate better recovery. The magenta values on the left represent the mean correlation for each algorithm, providing an overall performance metric.

In the moderate overlap scenario, most algorithms performed well, with the proposed algorithms (G, H, I), PCA+ICA, and PMD achieving particularly high correlations, indicating accurate recovery.  However, in situations with significant overlap, performance declined for some methods, especially PCA+ICA, PPCA+ICA, and SPCA+ICA which struggled to recover ground truth accurately. This suggested that combining PCA with ICA may not be sufficient for such complex scenarios, and SPCA+ICA was even worse than PCA+ICA, thus supporting the proposed hypothesis. GPower also showed lower correlation values, while PMD performed better but still fell short of the proposed algorithms.

\begin{figure}[t]
\includegraphics[width=7cm]{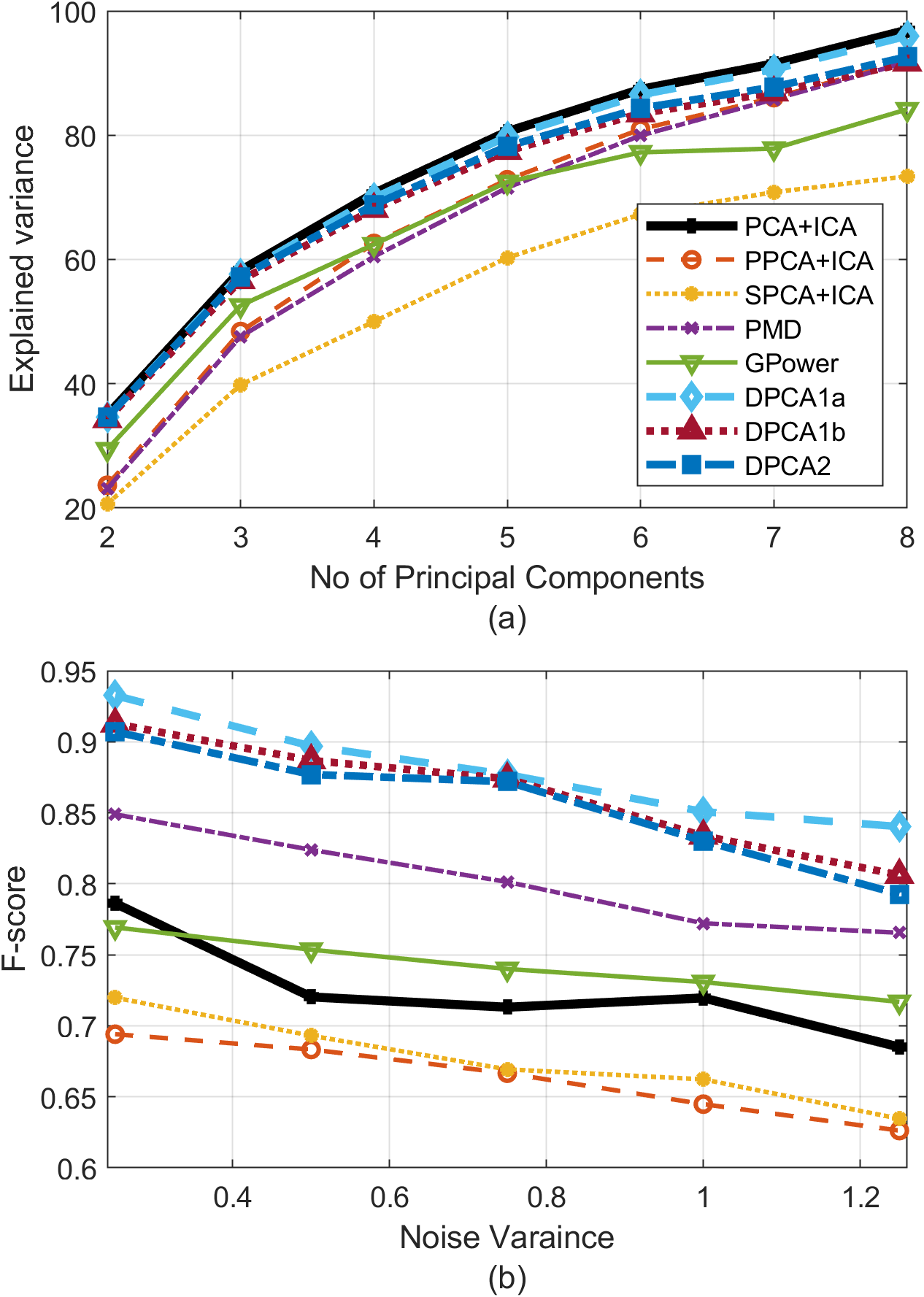}
\centering
\caption{a) Comparison of all algorithms on synthetic data to illustrate the variance explained as the number of principal components varied from $2$ to $8$, and b) mean F-score over $10$ trials calculated for various noise variances across all algorithms. \label{f2}}
\vspace{-3mm}
\end{figure}

\begin{figure}[t]
\includegraphics[width=7cm]{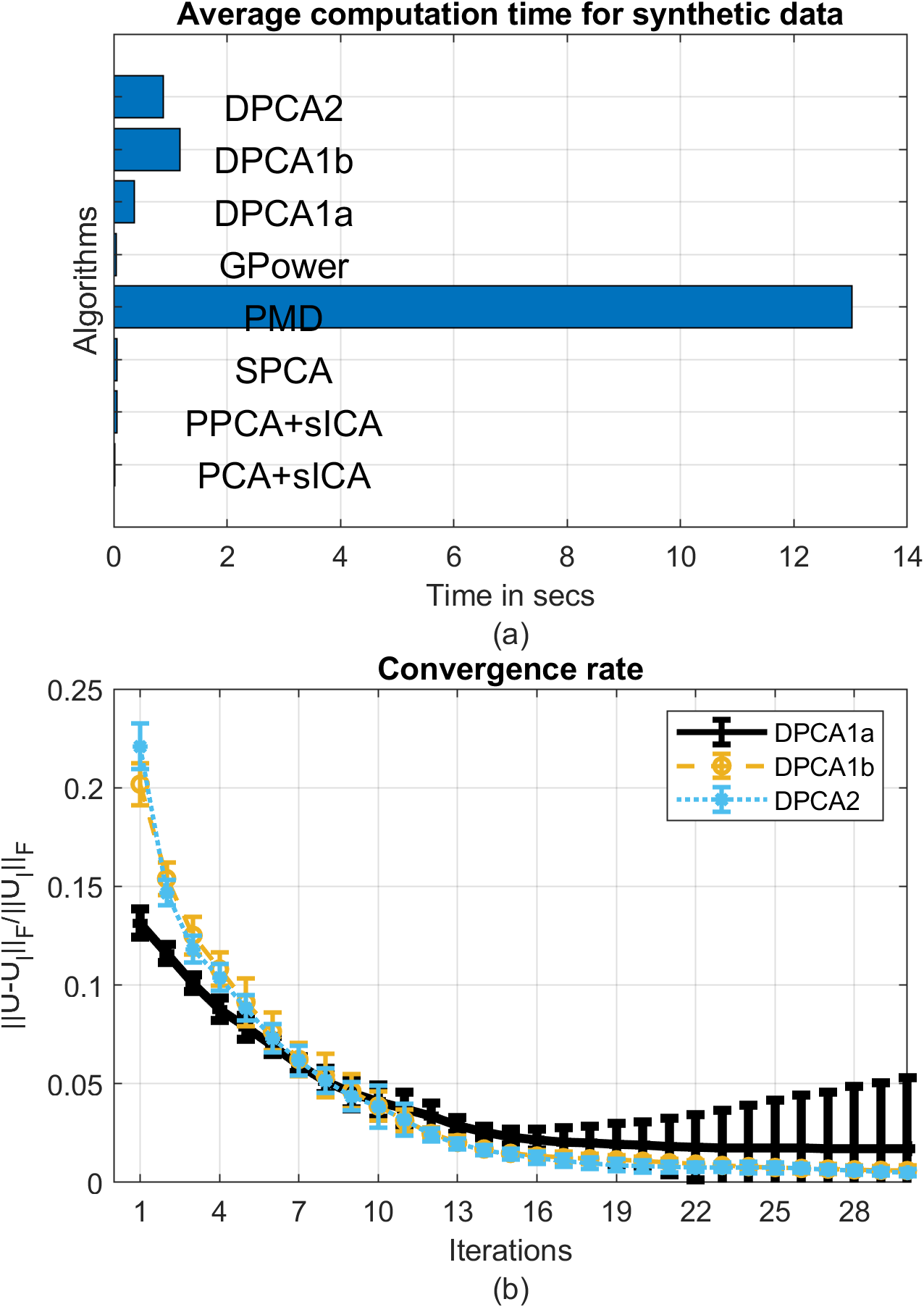}
\centering
\caption{a) Average computation time illustrating the time efficiency of various algorithms when processing synthetic data, and b) convergence rate displaying the convergence behavior of the DPCA variants, measured by the normalized difference between successive updates along with error bars indicating variability across runs. \label{f3}}
\vspace{-3mm}
\end{figure}

For each algorithm, the experiment from Figure $\ref{f1}$ was repeated $15$ times with temporal noise variance set to $0.9$ and spatial noise variance set to $0.005$, and spread parameter set to $6$. The results in Table \ref{T4} show the mean correlation values for each loading vector ($lv_1 - lv_8$), along with the overall mean and standard deviation for each algorithm. PPCA+ICA and SPCA+ICA produced the lowest mean correlations compared to other algorithms, while PCA+ICA and GPower achieved slightly higher means but with greater variability. PMD performed better, with a mean correlation of $0.920$ and a moderate standard deviation of $0.026$. The DPCA variants (DPCA1b and DPCA2) stood out, showing the highest mean correlations and lowest standard deviations, indicating superior consistency and accuracy. This indicates that DPCA1b and DPCA2 excel in accurately recovering GT-LVs, showing superior performance compared to traditional combinations of PCA and ICA.

Figure \ref{f2} compares several algorithms on synthetic data, focusing on explained variance and mean F-score under varying noise conditions. In subfigure (a), the explained variance is shown, as the number of PCs increased from $2$ to $8$, with temporal and spatial noise variances set to $0.3$ and $0.005$, and a spread parameter of $8$. The DPCA variants (DPCA1a, DPCA1b, and DPCA2) collectively achieved the second highest explained variance across all component settings, demonstrating their superior ability to capture data variance. GPower, and PMD showed moderate performance, while PCA+ICA and DPCA1a reached the highest explained variance levels among all algorithms.

Figure $\ref{f2}$b presents the mean F-scores over $10$ trials for each algorithm across varying noise levels, with temporal noise ranging from $0.25$ to $1.2$ and spatial noise from $0.001$ to $0.017$, while spread parameter set to $8$. The DPCA1a consistently achieved the highest F-scores, while DPCA1b and DPCA2 followed closely, demonstrating robust and accurate source recovery even in high noise environments. In contrast, the combination methods SPCA+ICA and PPCA+ICA displayed a rapid decline in F-scores as noise increased, signaling decreased effectiveness. Meanwhile, PMD and GPower exhibited a moderate decline but generally lower F-scores, underscoring less robustness compared to the DPCA variants.

Figure $\ref{f3}$ illustrates a comprehensive performance comparison of various decomposition algorithms under different noise conditions, extracted from a scenario where temporal and spatial noise levels were systematically varied as shown in Figure $\ref{f2}$b. In subfigure (a), the average computation time for each algorithm is charted, revealing the DPCA variants, particularly DPCA1b, as the most computationally demanding. Figure $\ref{f3}$b depicts the convergence rates of the DPCA variants. Each line indicates a consistent decrease in error as the iterations progress, which highlights the effectiveness of these algorithms (DPCA1b and DPCA2) in converging to a stable solution efficiently. Error bars indicate variability in convergence rates across trials, highlighting the stability of each algorithm under various data conditions. These insights reveal that DPCA1a experiences issues with convergence.

\section{Image Applications}
\subsection{Background subtraction on video sequences}
\begin{table*}[t]
\centering
\caption{For all algorithms involved in background subtraction, the sparsity parameters and computation time in seconds are reported as ${\lambda, \rho_1, \rho_2, \text{seconds}}$. The lowest computation time is highlighted in bold.}
\vspace{-1mm}
\begin{tabular}{l | c c c c c c}\toprule
\multirow{1}{*}{Dataset} & \multicolumn{1}{c}{SPCA} & \multicolumn{1}{c}{PMD} & \multicolumn{1}{c}{GPower} & \multicolumn{1}{c}{DPCA1a} & \multicolumn{1}{c}{DPCA1b} & \multicolumn{1}{c}{DPCA2}
\\\cmidrule(lr){1-7}
Highway traffic            & 300,4.5 & 0.45,\bf2.1 & 0.3,2.3  & 190,11.4 & 190,4,16,5.3 & 190,4,16,15.7\\
People in a corridor       & 250,1.7 & 0.45,1.3 & 0.01,2.4 & 600,3.8 & 600,12,32,\bf1.2 & 1600,12,32,8.2\\
Person crossing a street   & 390,5.0 & 0.6,2.9  & 0.1,\bf2.3  & 60,27.2 & 60,3,9,4.5 & 60,3,9,14.3\\\bottomrule
\end{tabular}
\label{T5}
\end{table*}
\begin{figure*}[t]
\begin{center}
\includegraphics[width=15cm]{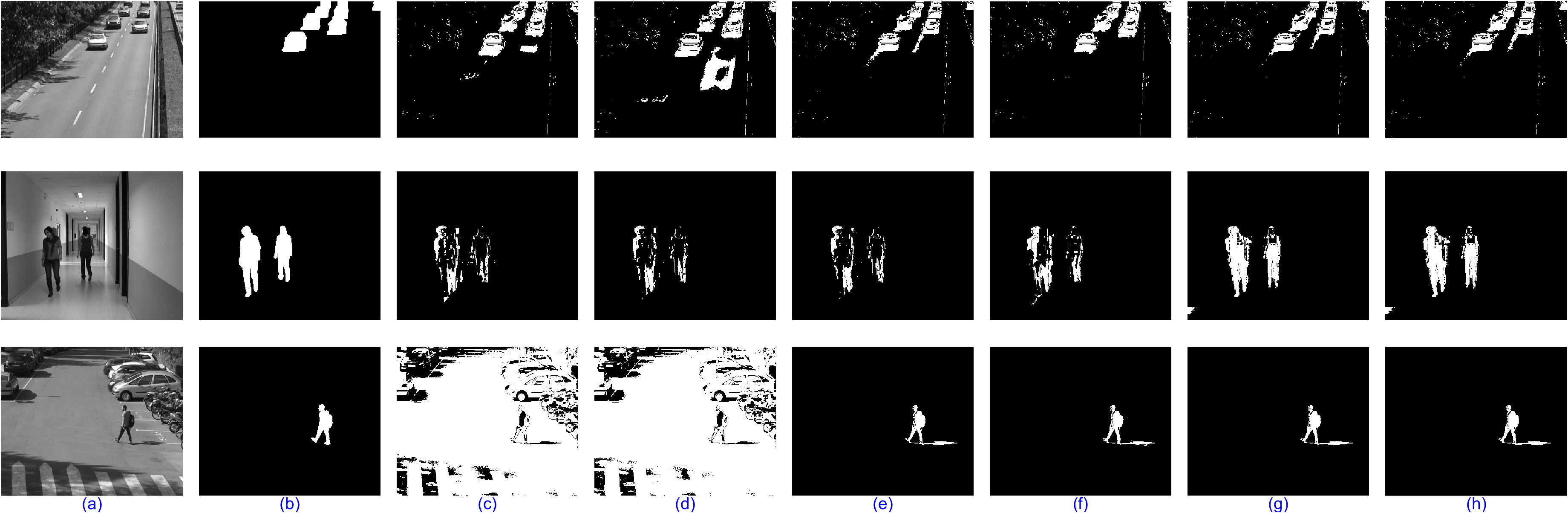}
\vspace{-2mm}
\caption{a) shows the original images from three distinct scenarios: highway traffic, people in a corridor, and a person crossing a street. b) displays the ground truth for foreground detection, (c) through (i) display the foregrounds recovered by various algorithms in the following order: SPCA, PMD, GPower, DPCA1a, DPCA1b, and DPCA2, each corresponding to the scenarios depicted.}
\vspace{-3mm}
\label{f4}
\end{center}
\end{figure*}

In this section, the performance of the proposed algorithms was evaluated through the application of background subtraction. This technique processed a video sequence, utilizing the frames to train ten PCs and their corresponding LVs which effectively model the background. Subsequently, pixels that significantly deviate from this decomposition model are identified as foreground. This method allows for dynamic separation of background and foreground elements, crucial for various video processing applications. This problem has been formulated as a sparse signal recovery problem in prior studies \cite{huangLearning2009, luOnline2013a, iqbalRobust2019, mirirekavandiLearning2024}, employing a data decomposition model denoted as $\text{X} = \text{U}\text{Z}$. This model facilitates the decomposition of vectorized video sequences stored in $\text{X}$ into PCs represented by matrix $\text{U}$ and their corresponding LVs contained within $\text{Z}$. After extracting the LVs, the background for each frame $\tilde{\text{x}}_i$ is reconstructed using a linear combination of PCs from the matrix $\text{U}$ and the sparse loading vector $\text{z}_i$, as $\tilde{\text{x}}_i=\text{U}\text{z}_i$. The foreground for each frame is then determined by calculating the residual $\text{x}_i - \tilde{\text{x}}_i$.

Three grayscale video sequences were considered from two different databases to make the evaluation diverse. This included a highway sequence from CDnet-2014 dataset\footnote{\url{https://www.kaggle.com/datasets/maamri95/cdnet2014}}, a bootstrap sequence and a person crossing a street from LASIESTA dataset\footnote{\url{https://www.gti.ssr.upm.es/data/lasiesta_database}}. The first sequence is of highway traffic that displays a fair number of foreground objects, appearing as medium traffic density in some frames and light traffic in others. The second sequence features two foreground objects consisting of two people walking in a corridor who shake hands in some frames and walk in opposite directions in others. The third sequence showcases a person crossing a street in a sunny area, while in later frames, another person crosses in the opposite direction in a shaded area.

The highway traffic video sequence has dimensions of $240 \times 320$ pixels across $1700$ frames. For evaluation purposes, only a subset of $400$ frames (from frame $700$ to $1099$) was utilized. The video of people in the corridor has dimensions of $288 \times 352$ pixels over $275$ frames, and the video of a person crossing the street comprises $288 \times 352$ pixels across $400$ frames. In all sequences, the camera position was static. Each frame was vectorized to form the columns of the data matrix $\text{X}$.

Only algorithms that demonstrated competitive performance were selected for this study, specifically SPCA, PMD, GPower, and the newly proposed algorithms. Each algorithm was configured to extract $10$ PCs, and the proposed algorithms were iterated for $20$ cycles. Default iteration settings were applied to the other algorithms. The optimal sparsity parameter for each algorithm was determined by experimenting with various values and selecting the one that yielded the highest F-score. Details of these sparsity parameters and the computation times for each algorithm are provided in Table \ref{T5}.

After extracting components, a query image from each video sequence was selected to retrieve its corresponding trained foreground and background. The original query images, their ground truths, and the thresholded recovered foregrounds are presented in Figure $\ref{f4}$. Overall, DPCA1a, DPCA1b, DPCA2, and GPower demonstrate more precise and accurate foreground detections across varied scenarios compared to SPCA and PMD. These latter algorithms struggle with consistency in complex scenes that feature multiple elements and variable lighting. Specifically, the DPCA variants excelled by accurately capturing most vehicles in the first dataset, providing clear and distinct outlines of individuals in the second dataset, and precisely capturing the silhouette of the person in the third dataset.

Table $\ref{T6}$, quantitatively evaluates the performance of six distinct algorithms by measuring precision, recall, and F-score across three varied scenarios. These results are correlated with the visual outcomes illustrated in Figure $\ref{f4}$. The data reveals that while SPCA and PMD provide useful insights, their performance fluctuates in complex settings. In contrast, the DPCA variants and GPower demonstrate consistently superior precision and F-scores, indicating not only higher accuracy but also robustness in handling diverse foreground extraction challenges effectively.

\begin{table*}[t]
\centering
\caption{Precision, Recall and F-score for the results shown in Figure $\ref{f4}$ using six different algorithms.}
\vspace{-1mm}
\begin{tabular}{l | c c c | c c c | c c c }\toprule
\multirow{3}{*}{Algorithm} & \multicolumn{3}{c}{Highway traffic} & \multicolumn{3}{c}{People in a corridor} & \multicolumn{3}{c}{Person crossing a street}
\\\cmidrule(lr){2-4}\cmidrule(lr){5-7} \cmidrule(lr){8-10}
           & Preci. & Recall & F-score  & Preci. & Recall & F-score  & Preci. & Recall & F-score  \\\midrule
SPCA & 0.725 & 0.700 & 0.712 & 0.541 & 0.239 & 0.331 & 0.012 & 0.667 & 0.023\\
PMD & 0.443 & 0.660 & 0.530 & 0.644 & 0.239 & 0.348 & 0.011 & 0.644 & 0.022\\
GPower & 0.804 & 0.677 & 0.735 & 0.685 & 0.256 & 0.372 & 0.780 & 0.757 & 0.768\\
DPCA1a & 0.807 & 0.713 & \bf0.757 & 0.733 & 0.368 & 0.490 & 0.785 & 0.756 & \bf0.771\\
DPCA1b & 0.830 & 0.692 &\bf0.755 & 0.828 & 0.776 & \bf0.801 & 0.782 & 0.756 & \bf0.770\\
DPCA2 & 0.817 & 0.691 & 0.749 & 0.825 & 0.778 & \bf0.803 & 0.782 & 0.757 & 0.769\\\bottomrule
\end{tabular}
\label{T6}
\vspace{-2mm}
\end{table*}
\begin{figure*}[t]
\begin{center}
\includegraphics[width=170mm]{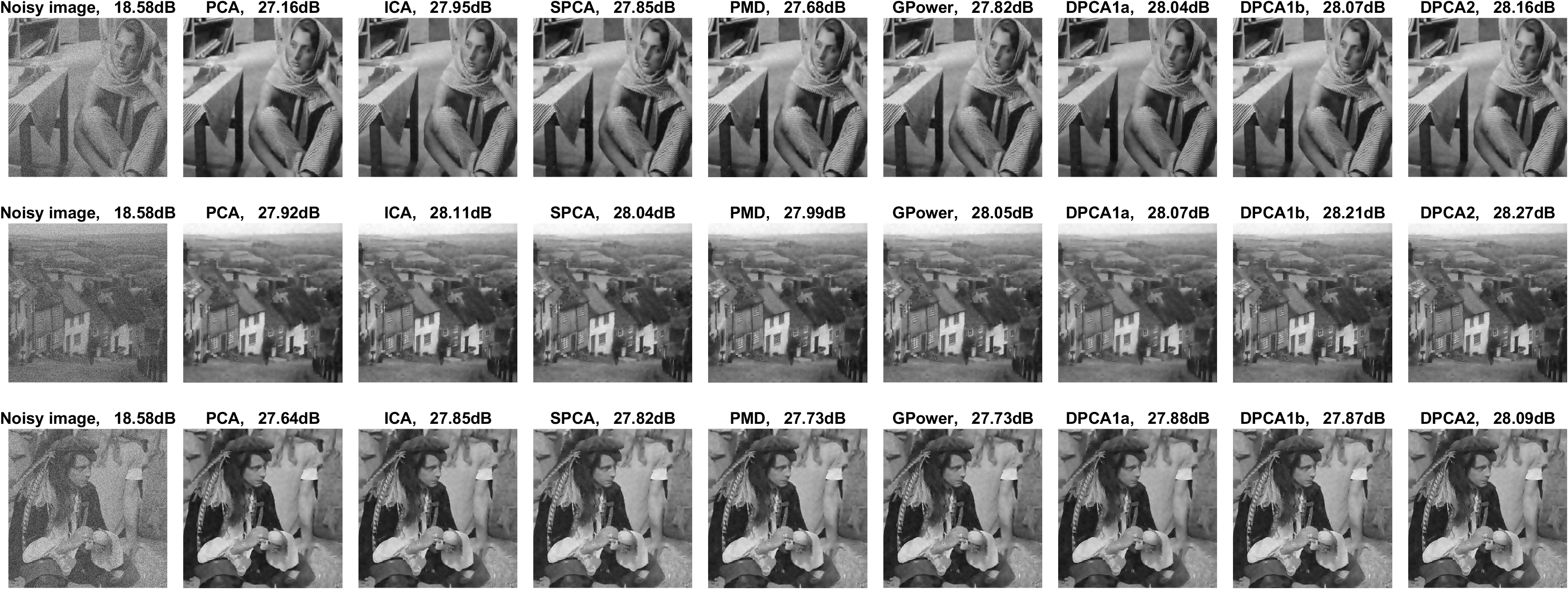}
\vspace{-2mm}
\caption{The first column display the original noisy images with an input PSNR of 18.58dB. Subsequent images illustrate the results of image reconstruction using eight different methods: PCA, ICA, SPCA, PMD, GPower, DPCA1a, DPCA1b, and DPCA2. Output PSNR values are noted above each image, demonstrating the effectiveness of each reconstruction technique across three distinct images: Barbara, Hill, and Man.}
\vspace{-3mm}
\label{f5}
\end{center}
\end{figure*}

\subsection{Image reconstruction}
This section evaluates the image reconstruction capabilities of the proposed algorithms against established methods including PCA, ICA, SPCA, PMD, and GPower. The test images were sourced from the miscellaneous category of the SIPI image database \cite{SIPI}. Each original image, sized $512 \times 512$, was converted to grayscale with pixel values ranging from $0$ to $255$. To simulate real-world conditions, Gaussian noise was added to these clean images. The noisy versions were then processed to create a training dataset. This was done by extracting $20000$ overlapping $8 \times 8$ patches using a sliding neighborhood approach, vectorizing these patches, removing their mean, and assembling them into the columns of the training matrix $\text{X} \in \mathbb{R}^{64 \times 20000}$ \cite{aharonKSVD2006, khalidEfficient2024}.

Only algorithms that demonstrated competitive performance were included in this study including PCA, ICA, SPCA, PMD, GPower, and the newly proposed DPCA variants. Each algorithm was configured to extract $64$ PCs, with the proposed algorithms iterating through $20$ cycles. Default settings were maintained for other algorithms. The optimal sparsity parameter for each was determined through experimentation, selecting values that maximized the PSNR, as detailed in Table $\ref{T7}$. After extracting the components, the denoised images were reconstructed by averaging pixels across all overlapping patches that had been processed for sparse representation \cite{iroftiPairwise2019}.

In Figure $\ref{f5}$, various image reconstruction techniques were applied to three different noisy images (Barbara, Hill, and Man), and their performance is quantified using the PSNR metric. PCA typically showed moderate improvement over the noisy input across all images. SPCA offered varying results, while it performed better on the Hill image, indicating some scenarios where the method may capture underlying structures effectively. ICA consistently provided a notable improvement in PSNR compared to PCA and SPCA. PMD displayed performance comparable to SPCA, highlighting its potential in scenarios where subtle textural features need to be preserved, but its computational inefficiency was quite high. GPower demonstrated strong performance across all images. DPCA variants (DPCA1a, DPCA1b, and DPCA2) consistently achieved the highest PSNR values among all algorithms. This indicates that these methods are particularly effective in handling noise and preserving or enhancing image details.

Table $\ref{T8}$ considers scenarios with even higher noise variance. It revealed that DPCA variants consistently outperform traditional methods like PCA, ICA, and SPCA across all images, demonstrating superior noise reduction capabilities while preserving image details. DPCA2 in particular often achieved the highest PSNR values, notably with the Hill and Man images, illustrating its effectiveness in managing various noise levels and image complexities. This highlights the advantages of integrating DMs and adaptive thresholding techniques in image reconstruction.

\begin{table*}[t]
\centering
\caption{For some of the algorithms involved in image reconstruction, the sparsity parameters are reported as ${\lambda, \rho_1, \rho_2}$.}
\vspace{-1mm}
\begin{tabular}{l | c | c c c c c c}\toprule
\multirow{1}{*}{Images} & \multicolumn{1}{c}{PSNRin} & \multicolumn{1}{c}{SPCA} & \multicolumn{1}{c}{PMD} & \multicolumn{1}{c}{GPower} & \multicolumn{1}{c}{DPCA1a} & \multicolumn{1}{c}{DPCA1b} & \multicolumn{1}{c}{DPCA2}
\\\cmidrule(lr){1-8}
Barbara   & 18.58 & 30 & 0.40 & 0.20 & 9000  & 9000,15,50  & 9000,15,50 \\
          & 14.14 & 40 & 0.45 & 0.15 & 50000 & 50000,10,45 & 50000,10,45 \\
          & 11.22 & 30 & 0.41 & 0.25 & 64000 & 64000,10,45 & 64000,10,45  \\\midrule
Hill      & 18.58 & 30 & 0.40 & 0.20 & 8000  & 9000,10,40  & 9000,10,40 \\
          & 14.14 & 40 & 0.45 & 0.15 & 32000 & 32000,10,45 & 32000,10,45\\
          & 11.22 & 30 & 0.41 & 0.25 & 64000 & 64000,10,45 & 64000,10,45 \\\midrule
Man       & 18.58 & 30 & 0.40 & 0.20 & 9000  & 9000,15,50  & 9000,15,50 \\
          & 14.14 & 40 & 0.45 & 0.15 & 32000 & 32000,10,45 & 32000,10,45\\
          & 11.22 & 30 & 0.41 & 0.25 & 64000 & 64000,10,45 & 64000,10,45\\\bottomrule
\end{tabular}
\label{T7}
\end{table*}

\subsection{Image inpainting}
In this section, the evaluation of the proposed algorithms is conducted through image inpainting tasks aimed at recovering missing pixels caused by superimposed text. The training dataset $\text{X}$ was composed of $1000$ randomly selected $8 \times 8$ patches from nine images sourced from the miscellaneous category of the SIPI image database \cite{SIPI}. All algorithms were then applied to this training set, which has dimensions of $64 \times 9000$, to extract PCs sized $64 \times 64$. Whereas two testing images were selected from this database, which were first painted using a text mask, and then divided into $N$ overlapping patches to produce an image matrix $\text{B}$ of size $64\times N$, where $N=62000$. Only algorithms that demonstrated competitive performance were selected for this study, specifically DCT bases of size $64\times 256$, PCA, DPCA1a, DPCA1b, DPCA2. All algorithms were run for $10$ iterations, and the best performing sparsity parameter was determined through extensive testing for each algorithm, and was found to be $\{\lambda,\rho_1,\rho_2\}=\{3000,10,40\}$ for the pepper's image and $\{3000,10,40\}$ for the couple's image.

\begin{figure*}[t]
\begin{center}
\includegraphics[width=14cm]{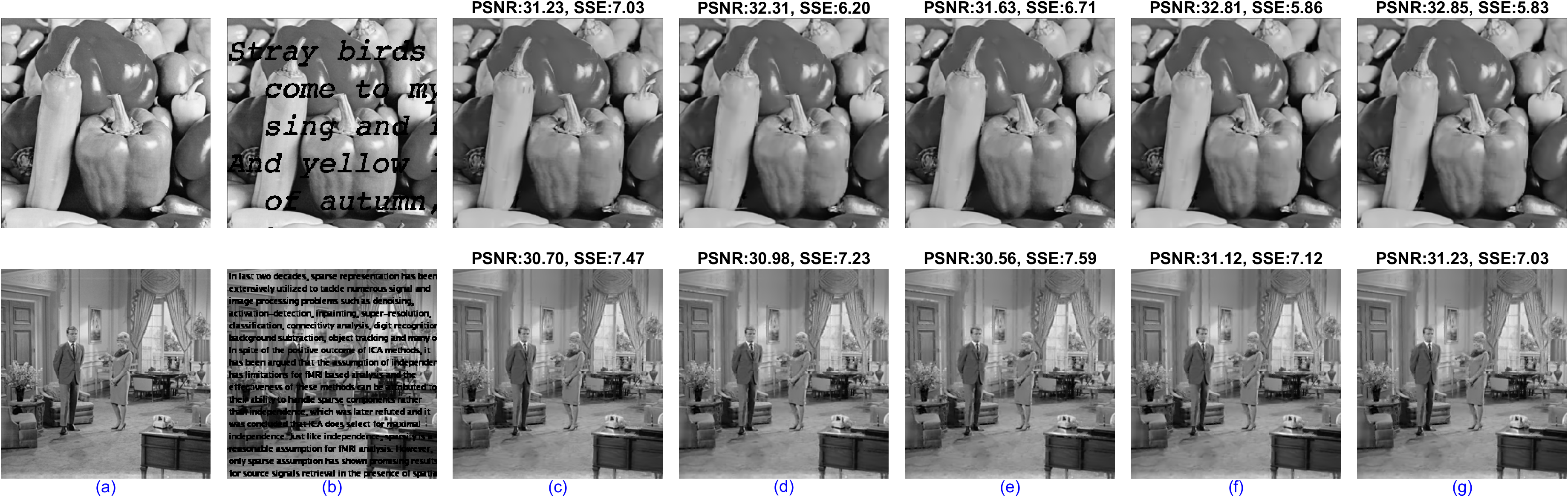}
\vspace{-2mm}
\caption{Image inpainting performance for two distinct scenarios: Peppers (top) and Couple (bottom). (a) shows the original images, (b) displays the images with text-overlay induced missing pixels, and (c) through (g) present the reconstructed images using DCT bases, PCA, DPCA1a, DPCA1b, and DPCA2, respectively. The performance of each reconstruction method is quantified by the PSNR and SSE values noted above each image.}
\vspace{-3mm}
\label{f6}
\end{center}
\end{figure*}

The Orthogonal Matching Pursuit (OMP) sparse coding technique was employed to learn sparse coefficients for each vector in $\text{B}$, aiming to minimize the reconstruction error $\norm{\text{M}\circ(\text{b}_i -\text{U}\text{z}_i)}_{F}^2$ for $i={1,...,N}$, where $\circ$ denotes element-wise multiplication, and $\text{M}$ represents the mask indicating missing pixels. The sparsity parameter was uniformly set to $20$ across all algorithms for both test images. The outcomes, displayed in Figure $\ref{f6}$, were quantified using the PSNR and the sum of squared errors (SSE) between the original and reconstructed images. Notably, the DPCA2 algorithm excelled in reconstructing the pepper image with minimal text overlay, achieving the highest image quality. Similarly, for the more challenging case with a greater number of missing pixels, DPCA2 again, outperformed all other algorithms.

In conclusion, the DPCA variants, particularly DPCA1b and DPCA2, demonstrate superior recovery capabilities in both scenarios, efficiently restoring images with substantial text-induced damage. These algorithms notably outperform traditional PCA and DCT bases, affirming the effectiveness of advanced sparse coding and integrating dissociation matrices in complex inpainting tasks.

\begin{table*}[t]
\centering
\caption{PSNR values in dB of noisy and denoised images recovered by seven different algorithms with best results highlighted in bold.}
\vspace{-1mm}
\begin{tabular}{l | c c c c c c c c}\toprule
Images & PSNRin  & PCA & ICA & SPCA & GPower & DPCA1a & DPCA1b & DPCA2 \\\midrule
Barbara&14.14 & 24.400 & 24.966 & 24.916 & 24.915 & 25.229 & \bf25.232 & 25.229\\
&11.22 & 22.871 & 23.102 & 23.136 & 23.130 & 23.331 & 23.362 & \bf23.386\\\midrule
Hill&14.14 & 26.087 & 26.164 & 26.165 & 26.180 & 26.246 & 26.214 & \bf26.282\\
&11.22 & 25.081 & 25.088 & 25.105 & 25.123 & 25.150 & 25.139 & \bf25.162\\\midrule
Man&14.14 & 25.582 & 25.663 & 25.675 & 25.643 & 25.772 & 25.788 & \bf25.906\\
&11.22 & 24.442 & 24.474 & 24.485 & 24.506 & 24.574 & 24.607 & \bf24.654\\\midrule
Mean& & 24.744 & 24.909 & 24.914 & 24.916 & 25.050 & 25.057 & \bf25.103\\\bottomrule
\end{tabular}
\label{T8}
\vspace{0mm}
\end{table*}

\section{Conclusion}
The proposed algorithms improve source separation and interpretability over traditional sparse PCA by estimating components jointly and modeling their interdependencies through dissociation matrices. Across the evaluated applications, the DPCA variants recover source structure more reliably than the compared sPCA and ICA pipelines, with the clearest gains under significant spatial overlap, although the method is not uniformly best on every metric and image. Future work will explore extensions across additional datasets and applications, potentially incorporating second level decompositions such as singular spectrum analysis \cite{hassaniSingular2022} and empirical mode decomposition \cite{kimEMD2009, khalidThree2024}.

\bibliographystyle{IEEEtran}
\bibliography{MyLibrary}

\end{document}